# Discrimination between Arabic and Latin from bilingual documents

Sofiene Haboubi, Samia Snoussi Maddouri, Hamid Amiri
System and Signal Processing Laboratory
National Engineering School of Tunis
BP 37 Belvedere, 1002, Tunis, Tunisia
sofiene.haboubi@istmt.rnu.tn; {samia.maddouri, hamid.amiri}@enit.rnu.tn

*Abstract*— **One of the important tasks in machine learning is the electronic reading of documents. The discrimination between languages is one of the first steps in the problem of automatic documents text recognition. We are interested in this work to the printed document Arabic and Latin (mixed). Our method is based essentially on the extraction of words from any document. Extracting structural features of each word; and then the recognition of language writing from a classification step. We present the found results of classification step, with a discussion on possible improvements.**

*Language identification;structural features; word extraction*

## I. INTRODUCTION

Discriminating between the languages a document image is a complex and challenging task. It has kept the scientists, working in this field, puzzled for quite some time now. Researchers have been emphasizing a lot of effort for pattern recognition since decades. Amongst the pattern recognition field Optical Character Recognition is the oldest sub field and has almost achieved a lot of success in the case of recognition of Monolingual Scripts.

One of the important tasks in machine learning is the electronic reading of documents. All documents can be converted to electronic form using a high performance Optical Character Recognizer (OCR). Recognition of bilingual documents can be approached by the recognition via script identification.

The digital processing of documents is a very varied field of research. Their goal is to make the machine able automatically to read the contents of a document, even of high complexity. Among the obtained results, we find the OCR (Optical characters recognition). This system makes to read the scripts form images, to convert them in numerical form. This system must know the language of script before the launching of process, to obtain good results. Also, the currently system does not treat two different languages in the same document. Like several researchers, our work is to introduce the Multi-language to our OCR, and we took the differentiation between the Arab and Latin scripts our field of contribution. Existing methods for differentiation between scripts are presented by [14] in four principal classes according to analyzed levels of information: methods based on an analysis of text block, methods based on the analysis of text line, methods based on the analysis of related objects and methods based on mixed analyses.

## II. BACKGROUND

The preliminary study, shown that the majority of differentiation methods treat only the printed text documents. Among the latter, the method suggested by [1], develops a strategy of discrimination between Arabic and Latin scripts. This approach is based on Template-Matching, which makes it possible to decide between the identified language. In [5], the authors uses a supervised Multi-Classes for classification and the Gabor filter to identify the Latin script. The type of document used is printed with Latin scripts and mixed. Two methods are proposed by [4], with two approaches: Statistical and spectral by Gabor filter. This work is interested on Kannada and Latin scripts. The system of identification, proposed by [13], relates to Latin and not-Latin languages in printed documents. This method is based on the application of Gabor filter, the author classifiers for the identification of languages other than Latin. With statistical methods and on printed documents, [15] has interested by identification of Arabic, Chinese, Latin, Devanagari, and Bangla languages. The identification of the type of scripts (printed or handwritten) is treated by [12], on Korean language. This approach is based on an analysis of related components and contours. A spectral method is presented in [16]. This method is to classes the script to Chinese, Japanese, Korean or Latin language by Gabor filter. The Arabic script is cursive and present various diacritic.

An Arab word is a sequence of letters entirely disjoined and related entities. Contrary to the Latin script, the Arab characters are script from the right to left, and do not comprise capital letters. The characters form varies according to their position in the word: initial, median, final and insulated. In the case of handwritten, the characters, Arabic or Latin, can vary in their static and dynamic properties. The static variations relate to the size and the form, while the dynamic variations relate to the number of diacritic segments and their order. Theirs increases the complexity of distinction between the languages.

## III. DISCRIMINATION BETWEEN LANGUAGES

The propose method consists of several stages (Figure 1). From a printed document, we must pass a filter to remove the diacritics dots. Then, extraction lines by horizontal projection. In the next step, we extract the words from lines; where each word will be treated separately.

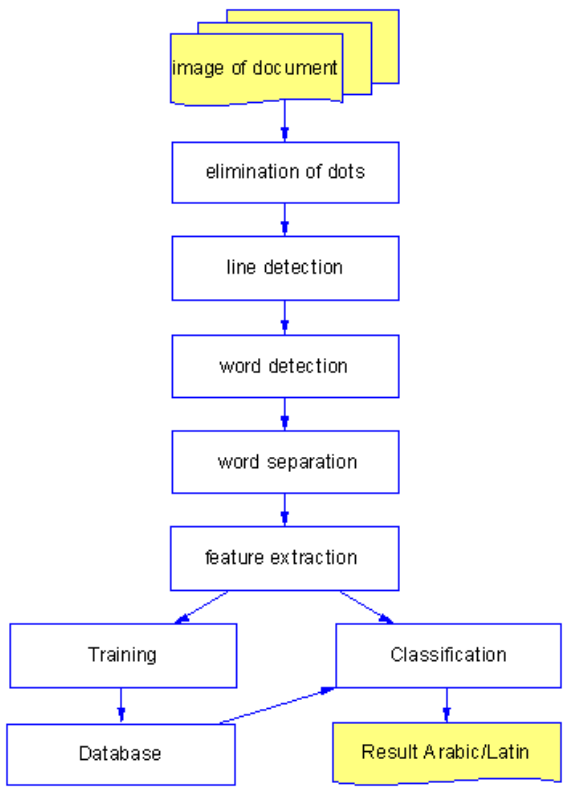

Figure 1. System architecture

We extracted the structural features for each word, and then passed to a stage of training or classification. In the training phase, we store all the results into a database, and in the classification phase, we use a classifier to determine the nature of language (Arabic or Latin).

*A. Elimination of dots*

The challenge is the presence of eight different types of diacritical marks, used to represent Arabic or Latin letters. In written text they are considered as special letters where each one is assigned a single code, as with normal letters. In fully diacriticized text a diacritical mark is added after each consonant of the Arabic word. These diacritical marks play a very important role in fixing the meaning of words.

But these diacritics can cause problems in the detection and extraction lines. Because in some writing styles, the diacritics may leave the upper limit of the line or the lower limit. This can cause poor detection of lines. Their diacritics will be removed by morphological erosion followed by a morphological dilation.

*B. Line detection*

To separate words from text document, we must first start with the detection of lines. There are several methods proposed by authors such as [7, 11]. But since we work with the Arabic and Latin, and our basic document that consists of text (without images and background), we chose to use the horizontal projection method. This method is still reliable in the absence of the inclination of lines. Figure 2 shows the application of the horizontal projection.

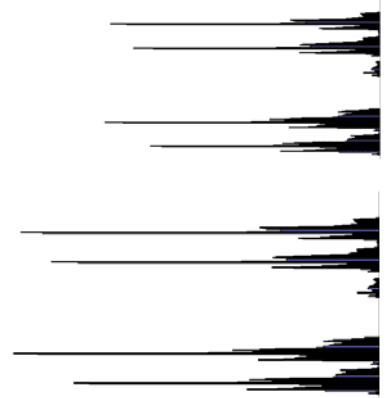

Figure 2. Line detection from Arabic/Latin Text

*C. Word detection and separation*

The detection and separation of words from a text document, is a conventional step in the automatic document processing. This step makes it difficult in the case of multilingual documents. This difficulty because of the large difference between the Arabic printed script and Latin printed script.

The Arabic handwriting is cursive. It is writing where the letters are linked to each other. This is not only in the case of handwriting; it is also the case of printed letters. The Latin alphabet is not cursive writing in the case of printed letters. In Arabic, the writing is always attached, even with print. This break is a problem in the step of separation between words; since both languages are present in the same text together.

Figure 3 shows the difference in dispersal areas in the case of Arabic (AR) and Latin (LA) scripts. To see this difference we used a Latin text and its similarity in Arabic, they have almost the same amount of information. We measured the distance (in pixels) between characters, and we calculated the number of occurrences of each distance found. In the case of the Arabic script, there are not much of separators between the characters, because of its nature cursive. In contrary, in the case of the Latin script, the distance is the dominant separators between the characters of the same word. But these two types of scripts have a common point at the threshold between the distances "between words" with the distances "between characters" (e.g. 6 pixels).

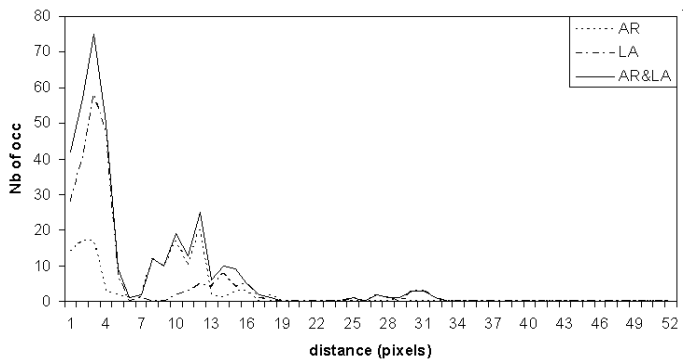

Figure 3. Dispersal spaces between characters and words

Our idea of separation of words is to use the Morphological image analysis to know the limits of the word (lower, upper, right and left). The structuring element for morphological dilation is a horizontal line form. The size of this line is the threshold that separates the distances between words and the distances between characters. In Figure 4 we show the impact of the order of dilation (size of the structuring element) on the number of connected components. The search for the ideal size of dilation is difficult with Figure 4.

In our case (Arabic and Latin text printed), the sequential dilation causes a decrease in numbers of connected components thereof (the characters of same word stick), then there is a stabilization, then there is a second decrease (the words stick). The difference between the two phases of reductions is stabilization, which is shown in Figure 5. Stability is the first value where the standard deviation of the variation in the number of related components vanishes.

After choosing the size of the structuring element, and after dilation of the original image, we determine the boundaries of each word in the Arabic and Latin text. The figure 6 shows an example of results found.

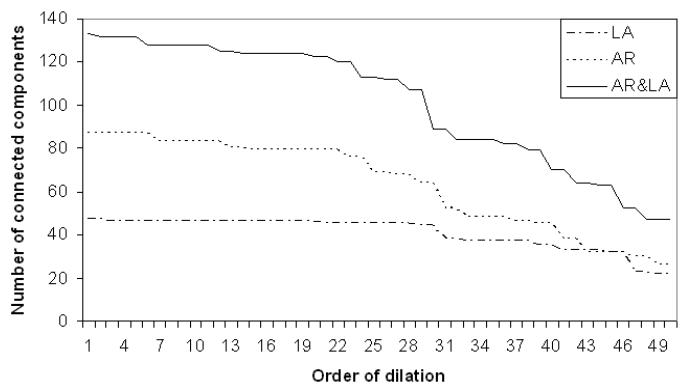

Figure 4. The impact of dialtion on the number of connected components.

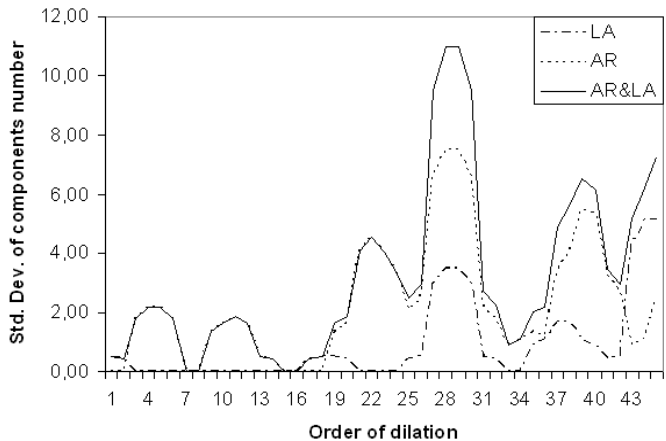

Figure 5. The standard deviation of the impact of dilation on the number of connected components.

The judge Al-Moaafi bin Zakariyya narrated from Ibrahim bin Faz'l, from Faz'l bin Yousuf, from Hasan bin Saber, from Wakee', from Hisham bin Urwah, from his father, from Ayesha, who said:

The Messenger of Allah ﷺ said, "Mentioning Ali bin Abi Taleb is worshipping Allah.

حدثنا أبو القاسم جعفر بن مسرور اللحام رحمه الله قال حدثني الحسين بن محمد عن إبراهيم بن محمد عن بلال عن إبراهيم بن صالح الأنماطي عن عبد الصمد عن جعفر بن محمد عن أبيه عن علي بن الحسين عن أبيه قال:

سئل النبي صلى الله عليه وآله عن قوله تعالى طوبى لهُمْ وَ حُسْنُ مَآبٍ قال نزلت في أمير المؤمنين علي و طوبى شجرة في داره و هي في الفردوس ليس من أثمار دور الجنة شيء إلا و غصن منها فيها.

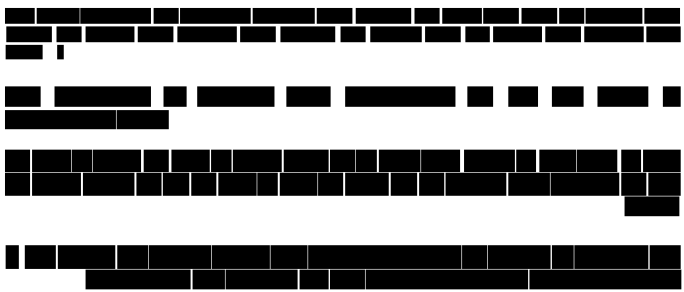

Figure 6. Distinction of words from bilingual text.

## D. Features extraction

The extraction of structural features is based on three steps: pre-treatment, the determination of the baseline, and the detection of primitives.

The words to be recognized are extracted from their contexts check or postal letters. A stage analysis, segmentation and filtering of documents is required. This task is not part of our work. The words are supposed to be taken out of context without noise. Since our method of feature extraction is based mainly on the outline, the preprocessing step we have introduced in our system is the expansion in order to obtain a closed contour with the least points of intersections. Since the elimination of the slope may introduce additional distortions we have tried to avoid this step. It is for this reason that techniques of preprocessing, avoiding the inclination correction has emerged [2, 3, 8, 11].

### 1) Determination of baselines

From the word we can extract two baselines. A upper and lower baseline. These two baselines divide the word into three regions. The poles "H" and diacritical dots high "P" which are regions above the upper baseline. The jambs "J" and diacritical dots lower "Q" correspond to regions below the lower baseline. The body of the word is the region between the two baselines. In general, the loops are in the middle.

### 2) Extraction of the poles and jambs

A pole is all forms with a maximum above the upper baseline. Similarly, jambs and all maxima below the lower baseline. The distance between these extrema and the baseline is determined empirically. It corresponds to:

*MargeH = 2(lower baseline – upper baseline) for the poles*

*MargeJ = (lower baseline – upper baseline) for the jambs.*

### 3) Detection of diacritical dots

The diacritical points are extracted from the contour. Browsing through it, we can detect those that are closed. From these closed contours, we choose those with a number of contour points below a certain threshold. This threshold is derived from a statistical study to recognize the words taken from their context (checks, letters, mailing ...).

### 4) Determination of loops from the contour

The problems encountered during the extraction of loops are: - Some diacritical dots can be confused with the loops if they are intersecting with the baselines. -Some loops may be longer than 60 pixels and can not be taken into account.

After a first selection step loops, a second step of verifying their inclusion in another closed loop is completed. This method involves: - Looking for word parts that can include the loop. - Stain the relevant section blank, if the contour points disappear when the latter is included in the word color and is part of the list of loops.

### 5) Detection of PAWS

Given the variability of the shape of characters according to their position, an Arabic word can be composed by more than one party called for PAW "Pieces of Arabic Word." Detection of PAWS is useful information both in the recognition step in the step of determining the position of structural features in the word.

### 6) Position detection primitives

The shape of an Arabic character depends on its position in the word. A character can have four different positions which depend on its position in the word. We can have single characters at the beginning, middle or end of a word. This position is detected during the primary feature extraction. Indeed, the extracted areas are defined by local minima. These minimums are from the vertical projection and contour. The number of black pixels is calculated in the vicinity of boundaries demarcated areas and between the two baselines above and below. If this number is greater than 0 at the left boundary and equal to 0 on the right, the position is the top "D", etc ...

## IV. TRAINING AND CLASSIFICATION

The recognition of the language of the document is regarded as a preprocessing step; this step has become difficult in the case of bilingual document. We begin by discriminating between an Arabic text and a Latin text printed by the structural method. Considering the visual difference between writing Arabic and Latin script, we have chosen to discriminate between them based on the general structure of each, and the number of occurrences of the structural characteristics mentioned above. Indeed, in analyzing a text in Arabic and Latin text we can distinguish a difference in the cursivity, the number of presence of diacritical dots and leg in the Arabic script. To printed Latin script, it is composed mainly of isolated letters.

The first step in the process regardless of the Arabic script from a text document is extracted lines and words. The extraction of lines is done by determining the upper and lower limit using the horizontal projection. For each line, there are the words using the method of dilation. Each word will be awarded by a system for extracting structural features.

From image of the word, we extract the feature vector that can discriminate between writing languages. This vector represents the number of occurrences of each feature in the word. It is to count the number of each feature: Hampe (H), Jamb (J), Upper (P) and Lower (Q) diacritic dots, Loop (B), Start (D), Middle (M), End (F) and Isolated (I). Our vector is composed of selected features

- the number of PAWS and the number of characters {NbPAW, NBL}
- the number of occurrences of each feature {H, J, B, P, Q, R, D, M, F , I}
- the number of occurrence of each feature taking into account their position {HD, HM, HF, HI, JF, JI, PD, PM, PF, PI, QD, QM, QF, QI, BD, BM, BF, BI}.

To evaluate our method, we used 57 documents Arabic and Latin text. After the step of separation into words, we found 4229 words. From each word, we generate its feature vector. For learning and testing, we used the Multilayer Perceptron function (MLP). The learning database contains 80% of all words, and the test database contains the rest. In practice, we used the WEKA[1] [6, 9] software to evaluate our method.

## V. RESULTS

In the test phase, we used 846 words (440 Arab and 406 Latin). After evaluating the test split, we found 798 words correctly classified and 48 words incorrectly classified. Who gave a classification rate equal to 94.32% and an error rate equal to 5.68%. From the confusion matrix, we found 37 Arabic words incorrectly classified and 11 Latin words incorrectly classified.

## VI. CONCLUSION

According to the method of discrimination presented, we showed that discrimination between Arabic and Latin is possible. The text documents used are in Arabic and Latin at the same time. The techniques used for discrimination are: mathematical morphology for separation into words, and structural characteristics for identification, and neural networks for classification. The levels found are too motivating to treat the case of handwritten documents.

---

[1] Weka: Data Mining Software. Weka is a collection of machine learning algorithms for data mining tasks. http://www.cs.waikato.ac.nz/ml/weka/